\documentclass[conference]{IEEEtran}

\usepackage{graphicx}
\usepackage{color}
\usepackage{placeins}
\usepackage{float}
\usepackage{tabularx,colortbl}
\usepackage{epstopdf}
\usepackage{amsmath}
\usepackage{amssymb}
\usepackage{caption}
\usepackage{subcaption}
\usepackage{algorithm}
\usepackage{algorithmic}
\usepackage{multirow}
\usepackage{url}

\usepackage{setspace}

\newcommand{\subfigsize}{.5}

\begin{document}

\title{Deep Learning Based Power Control for Quality-Driven Wireless Video Transmissions}
%
% Single address.
% ---------------
\author{Chuang Ye, M. Cenk Gursoy, and Senem Velipasalar
\\
Department of Electrical Engineering and Computer Science \\
Syracuse University \\
Syracuse, NY, 13244}

%\title{Deep Learning Based Power Control for Quality-Driven Wireless Video Transmissions}
%
%\author{Chuang Ye, M. Cenk Gursoy, and Senem Velipasalar
%\thanks{The authors are with the Department of Electrical
%Engineering and Computer Science, Syracuse University, Syracuse, NY, 13244
%(e-mail: chye@syr.edu, mcgursoy@syr.edu, svelipas@syr.edu).}}

\maketitle

\begin{abstract}
%\boldmath
In this paper, wireless video transmission to multiple users under total transmission power and minimum required video quality constraints is studied. In order to provide the desired performance levels to the end-users in real-time video transmissions while using the energy resources efficiently, we assume that power control is employed. Due to the presence of interference, determining the optimal power control is a non-convex problem but can be solved via monotonic optimization framework. However, monotonic optimization is an iterative algorithm and can often entail considerable computational complexity, making it not suitable for real-time applications. To address this, we propose a learning-based approach that treats the input and output of a resource allocation algorithm as an unknown nonlinear mapping and a deep neural network (DNN) is employed to learn this mapping. This learned mapping via DNN can provide the optimal power level quickly for given channel conditions.
\end{abstract}
\thispagestyle{empty}

% keywords

% \fi
%\IEEEpeerreviewmaketitle

\section{Introduction}
Recently, with rapid developments in communication technology, multimedia applications such as video telephony, teleconferencing, and video streaming which are delay sensitive and bandwidth intensive, have started becoming predominant in data transmission over wireless networks. For instance, as revealed in \cite{Cisco}, mobile video traffic accounted for 60\% of the total mobile data traffic in 2016, and more than three-fourths of the global mobile data traffic is expected to be video traffic by 2021. Indeed, mobile video has the highest growth rate of any application category measured among the mobile data traffic types. Such dramatic increase in wireless video traffic, coupled with the limited spectrum resources, brings a great challenge to today's wireless networks. Therefore, it is important to improve the wireless network capacity by allocating the limited resources efficiently.

The authors in \cite{Khalek} proposed a strategy to maximize the sum quality of the received reconstructed videos subject to different delay constraints at different users and a total bandwidth constraint in a multiuser setup by allocating the optimal amount of bandwidth to each user in a downlink wireless network. A content-aware framework for spectrum- and energy-efficient mobile association and resource allocation in wireless heterogeneous networks was proposed in \cite{Yiran}. The authors in \cite{Yichen} developed an optimal power allocation scheme for the cognitive network with the goal of maximizing the effective capacity of the secondary user link under constraints on the primary user's outage probability and secondary user's average and peak transmission power. The authors in \cite{Wenchi} proposed a QoS-driven power allocation scheme for full-duplex wireless links with the goal of maximizing the overall effective capacity under a given delay QoS constraint. Two models namely local transmit power related self-interference (LTPRS) model and local transmit power unrelated self-interference (LTPUS) were built to analyze the full-duplex transmission, respectively. However, an approximation of the sum Shannon capacity was used in the formulation of the effective capacity under the assumption that the signal-to-interference-plus-noise ratio is much larger than 1. \cite{YuWang} considered the problem of distributed power allocation in a full-duplex wireless network consisting of multiple pairs of nodes with the goal of maximizing the network-wide capacity. Shannon capacity was used as the performance metric and the optimal transmission powers for the full-duplex transmitters were derived based on the high SINR approximation and a more general approximation method for the logarithm function. In \cite{Chuang}, we addressed the optimal power and bandwidth allocation in a full-duplex wireless video transmission system with the goal of maximizing the weighted sum quality of the received video sequences.

The above-mentioned proposed algorithms achieve high performances that are observed through numerical simulations and theoretical analysis. However, the algorithms proposed for solving the optimal power control problems in the presence of interference terms generally have high computational complexity and cost. For example, WMMSE-based algorithms require complex computations such as matrix inversion and bisection in each iteration \cite{Shi}, \cite{Baligh}. Monotonic optimization (MO) in \cite{Chuang} also needs to find the projection in each iteration, in which one has to solve multiple nonlinear equations simultaneously and the candidate outline set increases. Such computationally demanding and time-consuming algorithms (due to iterative search processes as in monotonic optimization) become difficult to be implemented in real-time applications especially if wireless channel conditions vary relatively quickly, and hence optimal power levels need to be determined frequently.

In this work, we propose to employ a fully connected deep neural network (DNN) to approximate the optimal power control algorithm for quality-driven wireless multimedia transmissions. Since the optimization problem is not a convex problem, and solving the problem via monotonic optimization often entails high computational complexity and requires many iterations to converge even for a single parameter setting, the trained DNN model can provide the optimal power levels easily for different parameter settings and channel conditions.

%The remainder of this paper is organized as follows: The system model is presented in Section \ref{sec:System_Model}.   The optimization problems are formulated and the optimal policies are derived in Section \ref{sec:Maximization}. Simulation results are presented and discussed in Section \ref{sec:Result}. Finally, we conclude the paper in Section \ref{sec:Conclusion}.

\section{System Model} \label{sec:System_Model}
\begin{figure}
\centering
\includegraphics[width=0.5\textwidth]{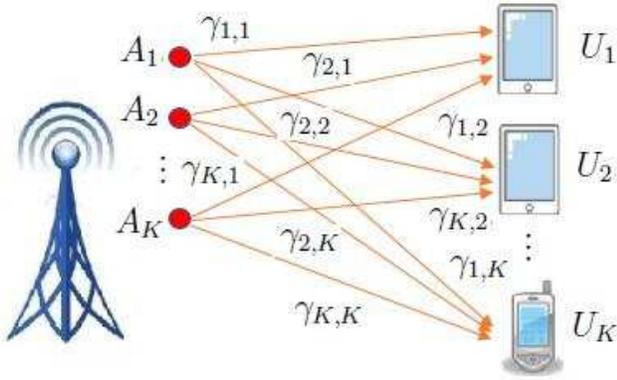}
\caption{Wireless system model in which each user receives multimedia data under quality and delay constraints.}\label{fig:System_Model}
\end{figure}

As depicted in Fig. \ref{fig:System_Model}, we consider a scenario in which a multi-antenna transmitter sends multimedia data to multiple users denoted by $U_k$ sharing the same spectrum with bandwidth $B$. There are in total $K$ users and the set $\mathcal{K} = \{1, 2, \ldots, K\}$ indicates the index of users.

%The practical application of this model includes, for instance, scenarios in which device-to-device (D2D) users exchange multimedia data (e.g., via social media sites) or conduct teleconferencing (i.e., engage in interactive video) in full-duplex mode. Assuming the availability of only statistical channel side information (CSI), base station acts as a coordinating agent and performs quality-driven resource allocation. Or in a different scenario, we can have one base station performing full-duplex multimedia communication with multiple users over different subchannels (e.g., via othogonal frequency division multiple access (OFDMA)). In this case, all the users on the left-hand side of Fig. \ref{fig:System_Model} essentially represent (or collapse to) a single base station in which there are multiple buffers and multiple flows of multimedia data to be sent to different users on the right-hand side. Base station again performs quality-driven resource allocation.

\section{Weighted Sum Quality-Maximizing} \label{sec:Maximization}

In this section, optimization problems are formulated to maximize the weighted sum video quality subject to maximum transmission power and minimum video quality constraints at each user. More specifically, we address the optimal allocation of transmission power levels assuming the availability of instantaneous CSI. It is assumed that all users share the same spectrum. Considering the adaptive modulation and coding (AMC) scheme adopted in the physical (PHY) layer, the transmission rate for $U_k$ can be rewritten as
\begin{align}
R_k = c_1 B \log_2\bigg(1 + \frac{P_k \gamma_{k, k}}{c_2(N_0 B + \sum_{i \neq k}^{K} P_i \gamma_{i, k} )}\bigg) \label{eq:tran_rate}
\end{align}
where $c_1$ and $c_2$ are rate adjustment and SNR gap, respectively \cite{Mazzotti}. $P_k$ is the transmission power level allocated to the transmission to user $k$, $\gamma_{i, j}$ is the channel fading coefficient between antenna $A_i$ and user $U_j$.

A PSNR-rate model for video in \cite{Luis} is employed to measure the quality of received video at user $U_k$, and the relationship between PSNR value and source rate is a logarithmic function that is expressed as follows:
\begin{align}
Q_k = \alpha_k \ln (R_k) + \beta_k. \label{PSNR_rate}
\end{align}
We can now express the weighted sum video quality for the proposed system as
\begin{align}
Q_{\text{tot}} = \sum_{k = 1}^K Q_k = \sum_{k = 1}^K \omega_k\big(\alpha_k \ln(R_k) + \beta_k\big), \label{weighted_sum_quality}
\end{align}
where $\omega_k \in [0, 1]$ denotes the weight for the quality of the video transmitted to user $U_k$ such that $\sum_{k=1}^K \omega_k = 1$. $Q_k$ is the quality of the received video at user $U_k$.

Now, the problem of maximizing the overall sum video quality of all users over power allocation strategies can be expressed as follows:
\begin{subequations}
\begin{align}
\max_{\substack{\mathbf{P} }} &\sum_{k = 1}^K \omega_k\big(\alpha_k \ln(R_k) + \beta_k\big) \label{Prob_Const1}\\
\text{s.t.}
& \sum_{k = 1}^{K} P_k \leq P^{\text{max}}; \quad  P_k \geq 0, \quad \forall k \in \mathcal{K} \label{Power_Const1} \\
& Q_k \geq Q_{k}^{\text{min}}, \quad \forall k \in \mathcal{K} \label{Quality_Const1}
\end{align}
\label{Prob_Const1}
\end{subequations}
Above, (\ref{Power_Const1}) is the maximum total transmission power constraint at the transmitter and (\ref{Quality_Const1}) is the minimum required video quality constraint. Specifically, $P^{\text{max}}$ and $Q_{k}^{\text{min}}$ are the maximum available transmission power at the transmitter and minimum received video quality at $U_k$, respectively. $\mathbf{P}$ is $K\times 1$ dimensional vector of power values, $[P_1, P_2, \ldots, P_K]$, allocated transmission to different users. The feasible sets of $\mathbf{P}$ is denoted by $\mathcal{P} = \{\mathbf{P}|\sum_{k = 1}^{K} P_k \leq P^{\text{max}}, P_k \geq 0, \forall k \in \mathcal{K}\}$.

\subsection{The Monotonic Optimization Algorithm}
Due to the existence of interference, the optimization problem (\ref{Prob_Const1}) is a non-convex problem with respect to $\mathbf{P}$. We note that the objective function in (\ref{Prob_Const1}) is an increasing function with respect to $\mathbf{R}$, where $\mathbf{R}$ is the $K\times 1$ dimensional vector of transmission rates, $[R_1, R_2, \ldots, R_K]$. Therefore, the non-convex optimization problem (\ref{Prob_Const1}) can be transformed into a monotonic optimization (MO) problem, and the MO formulation is written similarly as in \cite{Chuang} as follows:

\begin{align}
\max \Phi(\mathbf{R}) = &\sum_{k = 1}^K \omega_k\big(\alpha_k \ln(R_k) + \beta_k\big) \label{MO_fun1} \\
\text{s.t.} \quad & \mathbf{R} \in \mathcal{G} \cap \mathcal{H}.\label{eq:normal-conormal-sets}
\end{align}
Above, the normal set is
\begin{align}
\mathcal{G} = \left\{\mathbf{R}|0 \leq R_k \leq V_k(\mathbf{P}), \forall i \in \mathcal{I}, \forall k \in \mathcal{K}, \sum_{k = 1}^{K} P_k \leq P^{\text{max}} \right\}
\end{align}
where, $V_k(\mathbf{P})$ is the maximum feasible value of $R_k$.
In (\ref{eq:normal-conormal-sets}), the conormal set is
\begin{align}
\mathcal{H} = \{\mathbf{R}| R_k \geq R_k^{\text{min}}, \forall k \in \mathcal{K} \}
\end{align}
where $R_k^{\text{min}} = e^{\frac{Q_k^{\text{min}} - \beta_k}{\alpha_k}}$ is the minimum rate corresponding to the minimum quality of received video requirement.
Similar as in \cite{Chuang}, the initialized enclosing polyblock is obtained as the polyblock that contains the feasible set properly. In other words, the polyblock is the smallest box that contains $\mathcal{G} \cap \mathcal{H}$. We also initialize the optimal weighted sum quality of received video as $Q^{\text{opt}} = 0$. Following this, in each iteration $i$, we project the vertex $\mathbf{v}_i$ that leads to the maximum infeasible value of $\Phi(\mathbf{v}_i)$ to the upper boundary of the feasible set and get the corresponding feasible value of $\Phi(\mathbf{R})$, denoted as $\pi_{\mathcal{G}}^{\mathbf{u}}(\mathbf{v}_i)$, and $Q^{\text{opt}} = \max \{Q^{\text{opt}},  \pi_{\mathcal{G}}^{\mathbf{u}}(\mathbf{v}_i)\}$. After projection, we add $K$ new vertices to the polyblock set and remove the projected vertex $\mathbf{v}_i$. Repeating the above steps until $\Phi(\mathbf{v}_i) - Q^{\text{opt}} < \epsilon$ gives the optimal weighted sum quality of received video, where $\epsilon$ is the error tolerance. The corresponding $\mathbf{R}^{\text{opt}}$ is the optimal transmission rate, and we obtain the optimal power allocation $\mathbf{P}^{\text{opt}}$ by solving $K$ equations (\ref{eq:tran_rate}).

\subsection{Approximation via DNN}
In this section, we describe in detail the DNN structure as well as how the training and testing of the DNN are performed.
\subsubsection{DNN Structure}
\begin{figure}
\centering
\includegraphics[width=0.5\textwidth]{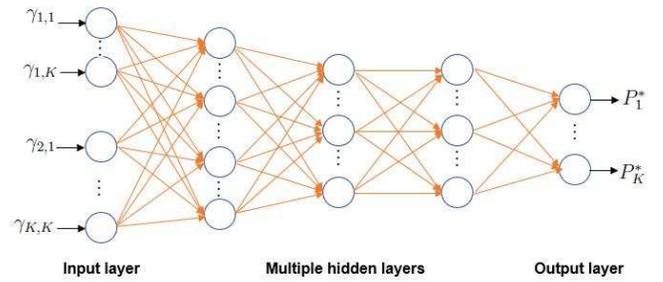}
\caption{The DNN structure used in this work.}\label{fig:DNN_Model}
\end{figure}

A fully connected neural network with one input layer, multiple hidden layers and one output layer as shown in Fig. \ref{fig:DNN_Model} is used in our paper. The input consists of the channel fading gains $\{\gamma_{i, k}\}$, and the output values are the power allocations $\{P_k\}$. Since the power value is greater than or equal to $0$, we employ ReLU as the activation function for the hidden layer activation, which gives the output of hidden layer $h = \max\{h, 0\}$. Additionally, in order to enforce the total power constraint in (\ref{Power_Const1}) at the output layer of DNN, the following normalized activation function for the output layer is used:
\vspace{-.3cm}
\begin{equation}
P_k = \frac{\max\{P_k, 0\}}{\sum_{k=1}^{K}\max\{P_k, 0\}} P^{\text{max}}.   \label{eq:normalize}
\end{equation}
The above equation guarantees that the total power constraint is satisfied with equality.

\subsubsection{Training Data Generation}
In order to describe the training data generation clearly, we define the vector $\mathbf{\gamma}^j = [\gamma_{1,1}^j, \gamma_{1,2}^j, \ldots, \gamma_{1, K}^j, \gamma_{2,1}^j, \ldots, \gamma_{K, K}^j]$ as the input of the training sample, and $\mathbf{P}^j = [P_1^j, P_2^j, \ldots, P_K^j]$ as the corresponding output (label), where the superscript $j$ denotes the index of the training sample. The corresponding power allocation, $\mathbf{P}^j$, is generated by using MO, and the tuple $(\mathbf{\gamma}^j, \mathbf{P}^j)$ is referred to as the $j$th training sample in the DNN structure. The MO process is repeated to generate the training set, as well as the validation data set. The size of the validation data set is smaller than the training set. $\mathcal{T}$ and $\mathcal{V}$ denote the sets of training and validation, respectively.

\subsubsection{DNN Training}
The training samples in $\mathcal{T}$ are used to optimize the weights and biases of the neural network. The mean square error (MSE) between the label $\mathbf{P}^j$ and the output of the DNN is used as the cost function. An efficient mini-batch stochastic gradient descent RMSProp algorithm that divides the gradient by a running average of its recent magnitude \cite{Hinton} is employed as the DNN optimization algorithm. The weights are initialized by using the truncated normal distribution.

\subsubsection{DNN Testing}
In the testing stage, the channel side information (CSI) is generated with the same distribution as used in the training stage. Each new generated channel realization $\mathbf{\gamma}$ is passed through the trained DNN and the optimal power allocation $\mathbf{P}$ is collected. Following this, we compare the power allocation and the corresponding weighted sum quality of received video sequences generated by DNN and MO.

\section{Numerical and Simulation Results} \label{sec:Resul}
Five CIF video sequences namely \emph{Akiyo, Bus, Coastguard, Foreman} and \emph{News} are used for the simulation results \cite{Reisslein}. Size of each frame is $352\times288$ pixels. FFMPEG is used for encoding the video sequences and group of pictures (GOP) is set as 10. Frame rate is set as 15 frames per second. Table \ref{table_qrate} shows the parameters $\alpha_k$ and $\beta_k$ that make the rate-distortion function of the five video sequences fit the quality rate model in (\ref{PSNR_rate}), where the unit of $R_k$ is kbit/s. Total bandwidth is $B = 10^5$ Hz. The power spectrum density of the AWGN is set to $N_0 = 10^{-6}$ W/Hz, and we also set the maximum available transmission power $P^{\text{max}} = K$ W, rate adjustment $c_1 = 0.905$ and SNR gap $c_2 = 1.34$ \cite{Mazzotti}.

\begin{table}[h]
\begin{center}
\caption{Parameter values of the quality rate model for different video sequences} \label{table_qrate}
{
\begin{tabular}{|c|c|c|c|c|c|}
\hline
  & Akiyo & Bus & Coastguard & Foreman & News \\ \hline
$a_k$ &  5.0545 & 4.7205 & 3.5261 & 4.5006 & 5.6218  \\  \hline
$b_k$ & 17.1145 & 5.4764 & 13.8425 & 13.0780 & 10.0016 \\  \hline
\end{tabular}}
\end{center}
\end{table}

\begin{figure}
\centering
\begin{subfigure}[b]{\subfigsize\textwidth}
\includegraphics[width=\textwidth]{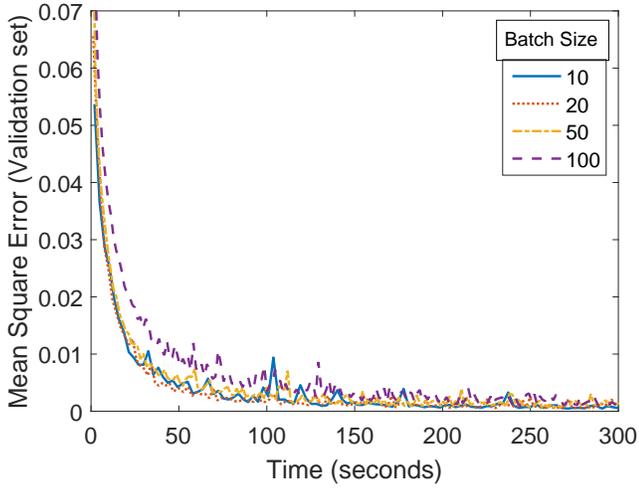}
\caption{\small{}}\label{fig:MSE_Bs}
\end{subfigure}
\begin{subfigure}[b]{\subfigsize\textwidth}
\includegraphics[width=\textwidth]{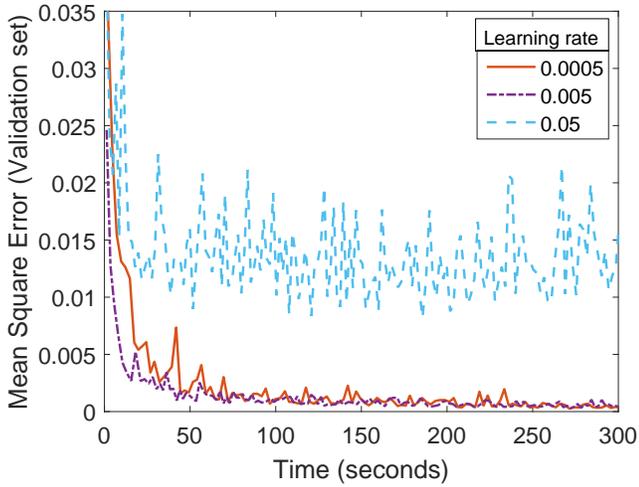}
\caption{\small{}}\label{fig:MSE_Lr}
\end{subfigure}
\caption{MSE are evaluated on the validation set. (a) Different curves represent different batch sizes (b) different curves represent different learning rates.}\label{fig:MSE_Bs_Lr}
\end{figure}

For the DNN, three hidden layers with $200$, $80$ and $80$ neurons in each hidden layer is used in our paper. We also assume that $K = 3$ and \emph{Akiyo, Bus, Coastguard} are transmitted from the base station to the users. Fig. \ref{fig:MSE_Bs} shows the impact of the batch size on the MSE evaluated on the validation set, as well as the training time with learning rate $0.0001$. Larger batch size leads to slower convergence of the training. Fig. \ref{fig:MSE_Lr} demonstrates the impact of the learning rate on the MSE with batch size $20$. Larger learning rate leads to higher validation error and may prevent convergence, while smaller learning rate results in slower convergence but lower error. And the validation error is at a certain level if the learning rate is small enough.

\begin{figure}
\centering
\includegraphics[width=\subfigsize\textwidth]{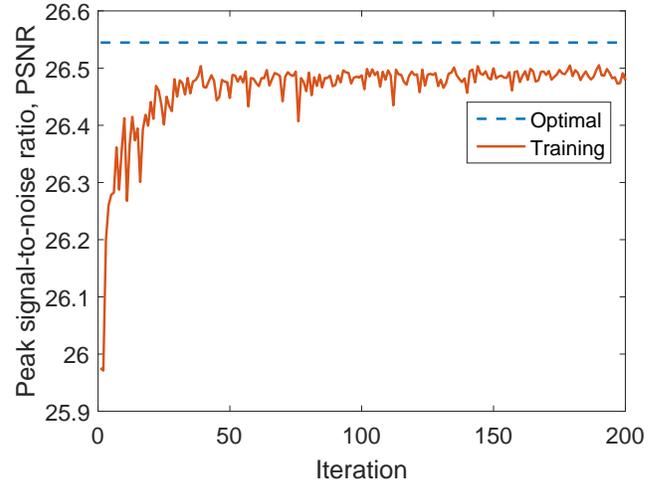}
\caption{Weighted sum quality of received video sequences in training stage.}\label{fig:PSNR_Traning}
\end{figure}

Fig. \ref{fig:PSNR_Traning} demonstrates the average PSNR value of all training samples as more iterations are performed. Learning rate is $0.0001$ and batch size is $20$. After around $30$ training iterations, the DNN converges and the average PSNR value fluctuates only very slightly around a certain value.

\begin{figure}
\centering
\includegraphics[width=\subfigsize\textwidth]{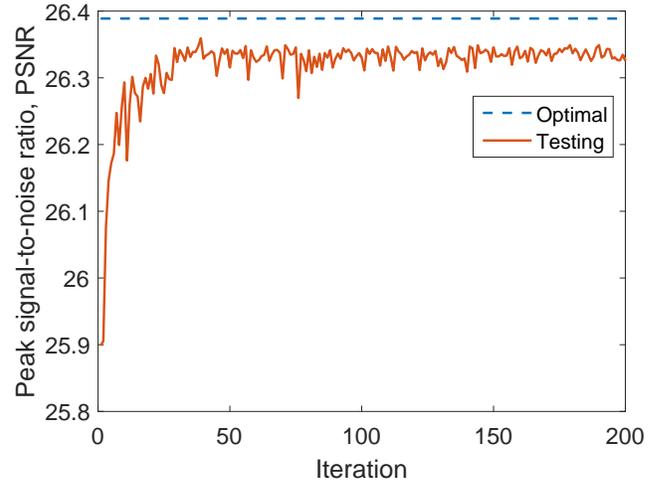}
\caption{Weighted sum quality of received video sequences in testing stage.}\label{fig:PSNR_Testing}
\end{figure}

\begin{figure}
\centering
\includegraphics[width=\subfigsize\textwidth]{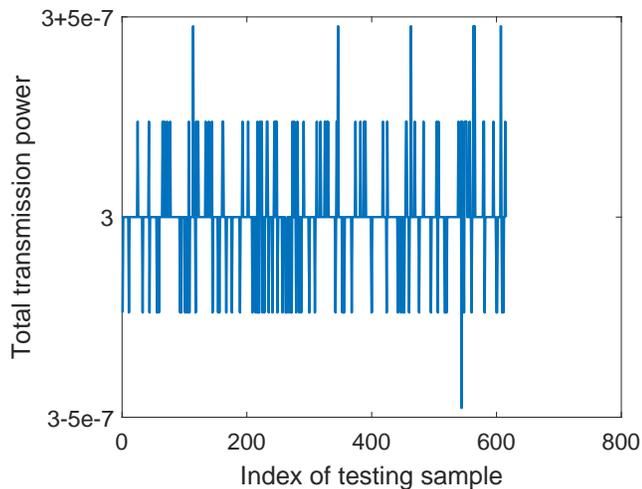}
\caption{Total power allocated to each sample in the testing set.}\label{fig:Power}
\end{figure}

Fig. \ref{fig:PSNR_Testing} demonstrates the average PSNR value of all testing samples again as more iterations are performed. The average PSNR value is much lower when the number of iterations is small, since the DNN is just trained a few times and that is not enough for getting the optimal DNN weights and biases. After around $30$ training iterations, the DNN converges and the average PSNR value for the testing set varies only slightly around a certain value that is a little lower than the optimal value (e.g., optimal PSNR level is around 26.4 dB while DNN provides approximately 26.35 dB).

After training the DNN $200$ times, Fig. \ref{fig:Power} plots the total power allocated to the testing samples. For each sample, the total power is around $3$ W, and the difference is less than $5 \times 10^{-7}$ W. Hence, the total power constraint is satisfied.

\section{Conclusion}
In this paper, we have addressed the maximization of the weighted sum quality of received video sequences under total transmission power constraint. We have reformulated the original nonconvex optimization problem as a monotonic optimization problem. The optimal power allocation levels that are generated via monotonic optimization algorithm is used as training set labels. We proposed a learning-based approach that treats the input and output of a resource allocation algorithm as an unknown nonlinear mapping and a deep neural network (DNN) is employed to train this mapping. This trained mapping can reduce the complexity and time consumption in determining the transmission power levels for given channel fading gains  while leading to very-close-to-optimal multimedia quality results measured in terms of PSNR.

\bibliographystyle{IEEEtran}
\bibliography{FullDuplex_DeepLearning}

\end{document}